\documentclass[letterpaper, 10 pt, conference]{ieeeconf}

\IEEEoverridecommandlockouts

\usepackage{etextools}
\usepackage{amssymb}
\usepackage{float}
\usepackage{makeidx}
\usepackage{amsmath}
\usepackage{amsfonts}
\usepackage{bbm}
\usepackage{epsfig}
\usepackage{epsf}
\usepackage{psfrag}
\usepackage{verbatim}
\usepackage{color}
\usepackage{multirow}
\usepackage{tabularx}
\usepackage{booktabs}
\usepackage[tight,footnotesize]{subfigure}
\usepackage{array}
\usepackage{soul}
\usepackage{footnote}
\usepackage{cite}
\usepackage{dblfloatfix}

\usepackage{color, colortbl}
\usepackage[colorlinks,bookmarksnumbered,citecolor=orange,urlcolor=orange]{hyperref}
\usepackage{graphicx}
\graphicspath{{Figures}}
\DeclareGraphicsExtensions{.pdf,.png}
\usepackage{bigstrut}
\usepackage[english]{babel}

\usepackage{csquotes}

\usepackage[T1]{fontenc}
\usepackage{algorithm, setspace}
\usepackage{algpseudocode}
\usepackage{url}
\usepackage{multirow}
\usepackage{float}
\usepackage{xcolor}
\usepackage{hyperref}
 \hypersetup{
     colorlinks=true,
     linkcolor=orange,
     filecolor=orange,
     citecolor=orange,      
     urlcolor=orange,
     }

\newcommand{\p}[1]{\smallskip \noindent \textbf{{#1}.}}
\newcommand{\eq}[1]{Equation~(\ref{eq:#1})}
\newcommand{\fig}[1]{Figure~\ref{fig:#1}}
\usepackage{balance}
\let\labelindent\relax
\usepackage{enumitem}
\begin{document}
\title{\LARGE
Prepare Before You Act: \\
Learning From Humans to Rearrange Initial States
}

%%%%%%%%%%%%%%%%%%%%%%%%%%%%%%%%%%%%%%%%%%%%%%%%%%%%%%%%%%%%%%%%%%%%%%%%%%%%%%%%
\author{Yinlong Dai, Andre Keyser, and Dylan P. Losey\vspace{-0.5em}
\thanks{This work is supported in part by NSF Grant $\#2337884$. The authors are members of the
Collaborative Robotics Lab (\href{https://collab.me.vt.edu/}{Collab}), Dept. of Mechanical Engineering, Virginia Tech, Blacksburg, VA 24061. \newline Corresponding author's email: \texttt{daiyinlong@vt.edu}}
}

%%%%%%%%%%%%%%%%%%%%%%%%%%%%%%%%%%%%%%%%%%%%%%%%%%%%%%%%%%%%%%%%%%%%%%%%%%%%%%%%

\maketitle
%%%%%%%%%%%%%%%%%%%%%%%%%%%%%%%%%%%%%%%%%%%%%%%%%%%%%%%%%%%%%%%%%%%%%%%%%%%%%%%%
\begin{figure*}[t]
	\begin{center}
        \includegraphics[width=\linewidth]{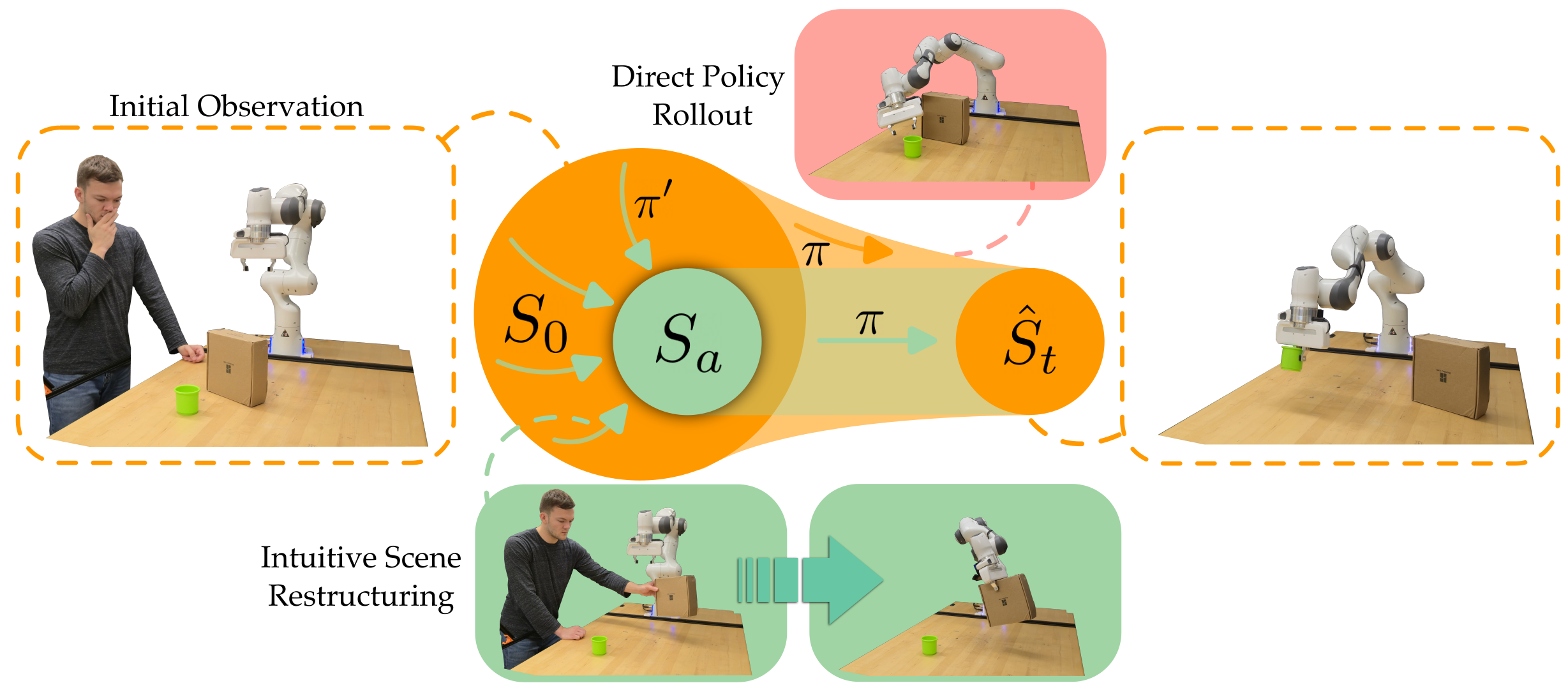}
		\caption{Robot arm learning to grasp a cup. When encountering an out-of-distribution initial observation $S_0$ (e.g., the cup is obstructed by a box), conventional approaches train on large-scale datasets and attempt to directly rollout the robot policy $\pi$. As we show, this brute-force approach falls short when the robot encounters unexpected initial environment states. By contrast, {ReSET} first learns a reduction policy $\pi^\prime$ based on how humans intuitively restructure the scene. This reduction policy rearranges objects (e.g., moving the box out of the way) so that the task is easier to perform and has lower state variance ($S_a$). Our approach then executes the default task policy $\pi$ from this simplified state distribution to reach the goal state $\hat S_t$.}
		\label{fig:front}
	\end{center}
    \vspace{-2.5em}
\end{figure*}

\begin{abstract}

Imitation learning (IL) has proven effective across a wide range of manipulation tasks. 
However, IL policies often struggle when faced with out-of-distribution observations; for instance, when the target object is in a previously unseen position or occluded by other objects. 
In these cases, extensive demonstrations are needed for current IL methods to reach robust and generalizable behaviors.
But when humans are faced with these sorts of atypical initial states, we often \textit{rearrange} the environment for more favorable task execution.
For example, a person might rotate a coffee cup so that it is easier to grasp the handle, or push a box out of the way so they can directly grasp their target object.
In this work we seek to equip robot learners with the same capability: \textit{enabling robots to prepare the environment before executing their given policy}.
We propose ReSET, an algorithm that takes initial states --- which are outside the policy's distribution --- and autonomously modifies object poses so that the restructured scene is similar to training data.
Theoretically, we show that this two step process (rearranging the environment before rolling out the given policy) reduces the generalization gap.
Practically, our ReSET algorithm combines action-agnostic human videos with task-agnostic teleoperation data to i) decide when to modify the scene, ii) predict what simplifying actions a human would take, and iii) map those predictions into robot action primitives.
Comparisons with diffusion policies, VLAs, and other baselines show that using ReSET to prepare the environment enables more robust task execution with equal amounts of total training data. See videos at our anonymous website: \url{https://reset2025paper.github.io}
                          
\end{abstract}

%%%%%%%%%%%%%%%%%%%%%%%%%%%%%%%%%%%%%%%%%%%%%%%%%%%%%%%%%%%%%%%%%%%%%%%%%%%%%%%%
\vspace{-0.5em}
\section{Introduction} \label{sec:intro}

Robots should be able to learn tasks from human demonstrations.
But learning seemingly simple manipulation tasks can become challenging under minor variations in the environment~\cite{sanchez2024recon, dai2025civil}.
Consider a setup in which an agent must grasp a cup (\fig{front}). 
A visuomotor policy can effectively complete this grasping task when the cup is directly observable. 
However, policies are prone to failure when facing states outside the training distribution; for example, if visual access to the object is obstructed by a box. 
As shown in \fig{front}, when we try to execute a direct policy rollout --- and the box is in the way --- the robot does not know what to do, leading to critical mistakes like missing the cup.

One standard approach is to try and overcome this challenge is by collecting larger and more diverse sets of training data.
Given enough examples, the robot can figure out how to grasp around the box~\cite{dalal2025local, bharadhwaj2024track2act}.
However, we hypothesize that this is not the most efficient or time-effective way to solve the problem, particularly for long-horizon tasks.
Instead, we propose an alternate approach based on how humans go about these inconvenient or unexpected initial states.
Imagine a person faced with the scenario in \fig{front} --- rather than attempting to reach around the box, humans often simplify the problem by first restructuring the environment. 
For instance, here a person could move the box out of the way, making the task of grasping the cup much easier to complete. 
Inspired by the way humans prepare environments, our insight is that:
\begin{center}\vspace{-0.4em}
\textit{We should design policies that restructure the environment, bringing diverse start states into a familiar and manageable distribution before executing the actual task.}
\vspace{-0.4em}
\end{center}
Applying this insight we introduce \textbf{ReSET}, \textbf{R}estructuring \textbf{S}tates for \textbf{E}fficient policy \textbf{T}raining.  
Rather than learning one policy that tries to complete the entire task across a broad range of initial states, under ReSET we learn two policies.
The first policy is a \textit{reduction policy} that simplifies the initial state by bringing it into a tighter distribution (e.g., moving objects out of the way).
The second policy --- the default \textit{task policy} --- then completes the task from that known distribution (e.g., picking up the cup).
By efficiently learning from human teachers how to restructure the environment, we effectively reduce task variability, enabling default policies to perform more complex manipulation tasks without requiring as much total training data.

Overall, we make the following contributions:

\p{Generalization and Data Efficiency}
We provide theoretical analysis that suggests learning a reduction policy on top of a task policy i) yields a lower generalization gap upper bound, and ii) requires less total training data.

\p{Flow-Based Approach}
We learn a flow-based reduction policy from action-agnostic human demonstrations and task-agnostic robot play data.
This approach i) determines when to rearrange objects, ii) predicts how objects should be manipulated, and iii) maps visual point flows to robot actions.

\p{Experimental Validation}
We show that ReSET outperforms several state-of-the-art baselines in handling out-of-distribution states across a range of few-shot task settings.

\section{Related Work} \label{sec:related}

\noindent \textbf{Imitation Learning.}
Recent advances in imitation learning enable robots to learn policies capable of executing long-horizon tasks in complex real-world settings \cite{hussein2017imitation, haldar2024baku, zitkovich2023rt}. 
Diffusion policies \cite{chi2024diffusionpolicy} model the distribution over actions using denoising diffusion probabilistic models, enabling multimodal behaviors.
Flow-matching approaches such as \(\pi_0\) \cite{black2024pi_0} learn a continuous velocity field that directly maps noise to actions. 
On the other hand, Vision-Language-Action (VLA) models \cite{zitkovich2023rt, kim24openvla} leverage the generalization capacity of pretrained large vision-language models, unifying vision, language, and actions within a shared feature space.
But regardless of whether we use diffusion policies, flow-matching approaches, or VLAs, the model must fundamentally be able to work across a diverse set of initial states (e.g., different object locations, clutter, and visual features).
However, few-shot deployment to scenarios outside of the training distribution remains an open challenge  \cite{gupta2025adapting, mao2025omnid}.

\p{Scaffolding Approaches}
There are a variety of recent IL works designed to handle out-of-distribution states.
We broadly refer to these strategies as ``scaffolding.''
One simple scaffolding approach is to collect diverse data.
Works such as \cite{lin2024data, mandlekar2019scaling,  ebert2021bridge} improve generalization by acquiring large-scale real-world datasets. 
But despite recent advances in data collection methods \cite{iyer2024open, zhao2023learning}, gathering millions of diverse trajectories still remains difficult and costly.

To work around the issue of manual data collection, methods like \cite{chen2024semantically, mandi2022cacti,ameperosa2024rocoda} leverage data augmentation or domain randomization and generate synthetic datasets. 
In practice, real world synthetic data can also be produced using generative models (e.g., diffusion models). 
However, careful prompting is required to avoid producing unrealistic or confusing samples \cite{chen2024semantically}, and ultimately there remains a gap between real data and synthetic images.

The methods most closely related to our proposed approach are object-centric recovery (\textit{OCR}) \cite{gao2024out} and dynamics-augmented diffusion policy (\textit{Dynamics-DP}) \cite{wu2025neural}.
\textit{OCR} assumes direct access to object positions and learns a recovery policy by exploiting gradients on the object keypoint manifold within the training data. \textit{Dynamics-DP} learns a dynamics model from robot play data in simulation and employs model-based control to generate augmented trajectories that capture recovery from out-of-distribution states.
In practice, we find that both of these approaches assume access to ground-truth states and require significantly more training data that our proposed method.
However, the central message conveyed by these works --- reshaping the environment into a regime more tractable for policy execution --- aligns with our research direction.
Building on this perspective, we next provide a theoretical analysis demonstrating that restructuring can improve the performance of learned policies.

\section{The Effects of Anchor States} \label{sec:problem}

Returning to our motivating example, consider a robot arm trying to grasp a cup occluded by a box (see \fig{front}). 
Existing approaches typically attempt to reason about out-of-distribution states and directly roll out the policy (i.e., maneuvering around the obstruction and grasping the object without visual access).
By contrast, ReSET transforms the initial state into a set of simpler, more tractable intermediate states which we call the $\textbf{anchor states}$.
In other words, our approach first reveals the cup by removing the box, and then proceeds to execute the given robot policy.

But is breaking the task into two parts and moving to these anchor states actually data efficient?
In this section we show that --- given the same amount of training data --- leveraging a set of anchor states lowers the theoretical upper bound on the policy’s generalization gap.
This enables the policy to perform successfully across a wider distribution of initial states. 
We begin by defining the the generalization gap we aim to bound and the anchor states we try to enforce.

\p{Definitions}
Let $s\in \mathbb{R}^d$ be the state of the environment. A robot policy $\pi: s \to u$ maps states to robot actions $u \in \mathbb{R}^m$. 
When executed in an environment with a transition kernel $P_\theta (s^\prime | s, u)$, the policy yields a Markov transition operator over states \cite{kaelbling1998planning}: 
\begin{equation}
T_{\pi, \theta} = \sum_u P_{\theta}(s^\prime | s,u) \pi (u | s)
\end{equation}
where $\theta \in \Theta$ is the environment dynamics. Starting from the initial state distribution $S_0$ and executing $t$ actions, the policy $\pi$ produces a distribution over trajectories through its interaction with the environment.
The induced distribution over final states can be expressed as
$S_t \sim T_{\pi, \theta}^{t}(S_0)$. 
Here $T_{\pi,\theta}^{t} \;=\; \prod_{k=1}^{t} T_{\pi,\theta}
$ denotes the $t$-fold composition of the Markov transition operator, conditioned on both the policy $\pi$ and the environment dynamics $\theta$. 

\p{Generalization Gap}
Following \cite{kawaguchi2023does}, we define the generalization gap of the $t$-fold Markov transition operator induced by policy $\pi$ as the difference between its expected loss under the true distribution and its empirical loss on the training set:  
\begin{equation} \label{eq:2}
\begin{split}
    \Delta(\mathcal{D}, S_0) \;=\; \mathbb{E}_{(S_0, \hat S_t)}\!\left[l\!\left(T_{\pi,\theta}^{t}(S_0), \hat S_t\right)\right] \;-\;  \\\frac{1}{n}\sum_{i=1}^{n} l\!\left(T_{\pi,\theta}^{t}(s_0^i),\, \hat s_t^i\right)
\end{split}
\end{equation}
where $\hat S_t$ denotes the desired goal state at time $t$, and the training dataset is given by 
$\mathcal{D} = \{(s_0^i, \hat s_t^i)\}_{i=1}^n$ containing $n$ samples drawn from the same distribution as the random variable pair $(S_0, \hat S_t)$.
Here $l(\cdot,\cdot)$ denotes a loss function measuring the discrepancy between the predicted state and the target state. 
This generalization gap indicates how well a policy will perform on new states. 
We want to constrain the generalization gap so that the policy performs just as well on unseen situations as it does on the training data.

\p{Anchor States}
We define \textit{anchor states} as $S_a \sim T_{\pi^\prime, \theta}^{a}(S_0)$, a set of states that the robot visits under a reduction policy $\pi'$ at timestep $a$ such that $0 < a < t$. 
Anchor states are a subset of states with a more concentrated distribution, making them more tractable for the base policy to execute trajectories from (see \fig{front}). Formally, we say that $S_a$ constitutes an anchor if its distribution is more concentrated than that of the initial state distribution $S_0$. We formalize this condition by requiring that the trace of the covariance matrix of $S_a$ satisfy $\operatorname{tr}(\Sigma_a) < \operatorname{tr}(\Sigma_0)$. Here $\Sigma = \operatorname{Cov}(S)$ is the covariance matrix of $S_t$ and the trace of the matrix is $\operatorname{tr}(\Sigma) = \sum_{i=1}^n \sigma_i^2$, where $\sigma_i^2$ denotes the variance of the $d$-dimensional state-space along the $i^{th}$ dimension.
In our motivating example, $S_a$ can represent the set of states where the box has been removed and the object is directly observable by the robot.
Substituting $S_a$ in \eq{2}, we can rewrite the expression for generalization gap as: 
\begin{equation} \label{eq:t2}
    \begin{split}
    \Delta(\mathcal{D}, S_a) \;=\; \mathbb{E}_{(S_0, \hat S_t)}\!\left[l\!\left(T_{\pi,\theta}^{t}(T_{\pi^\prime,\theta}^{a}(S_0)), \hat S_t\right)\right] \;-\;  \\\frac{1}{n}\sum_{i=1}^{n} l\!\left(T_{\pi,\theta}^{t}(T_{\pi^\prime,\theta}^{a}(s_0^i)),\, \hat s_t^i\right).
\end{split}
\end{equation}

\begin{figure*}[!t]
	\begin{center}
 		\includegraphics[width=\linewidth]{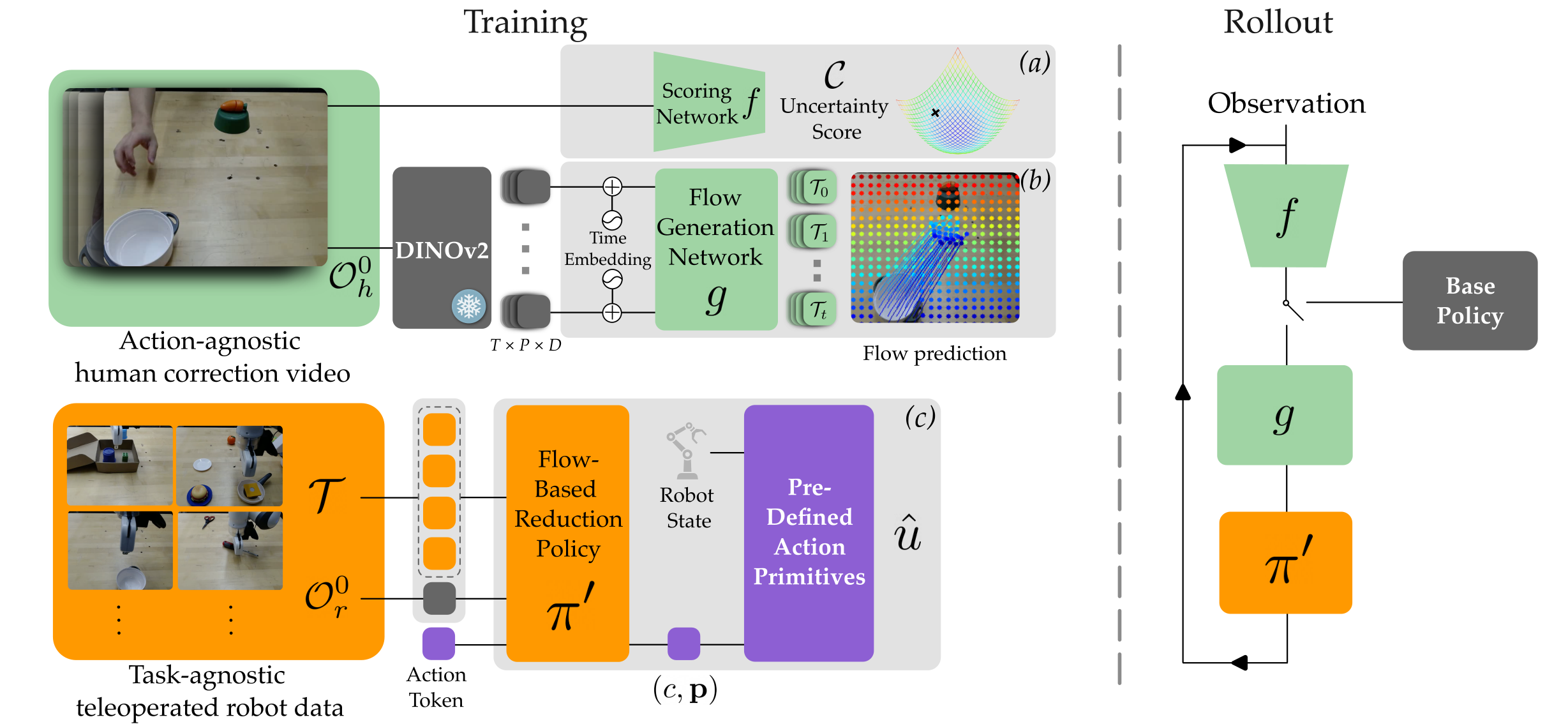}
        % SVG: https://drive.google.com/drive/u/3/folders/1pdDgtGy_qHmcoVWgfEvZOjG3BnNS4_jm
		\caption{\textit{Left}: Network architecture of ReSET. The model consists of three key components: (a) a scoring network $f$, trained on human videos, which estimates the likelihood that a base policy will succeed under a given initial configuration; (b) a flow generation network $g$, which predicts flows encoding human intuition about how the scene should be restructured into anchor states; and (c) a reduction policy $\pi^{\prime}$ that translates the predicted flows $\mathcal{T}$ into executable robot action primitives $\mathcal{A}$.
        \textit{Right}: At rollout, the scoring network evaluates the current observation to determine whether it is ready to execute the base policy. If not, the flow generation network produces a flow plan, and the reduction policy will execute that plan. The scoring network then re-evaluates the updated scene before deciding whether to proceed with the base policy.
        }
		\label{fig:method}
	\end{center}
    \vspace{-2.5em}
\end{figure*}

\subsection{Using Anchor States to Reduce Generalization Error}\label{AA}
Next we will show that constraining the state distribution by creating anchor states can lower the upper bound on the expected generalization gap, thus improving policy performance. 
We make the following assumptions in our analysis: i) the operator $T_{\pi^\prime, \theta}$ is linear and ii) the distribution $S$ is centered. Formally, we get the tightest upper bound on the generalization gap $\sup \, \mathbb{E}[\Delta(\mathcal{D}, S_a)] \;<\; \sup \, \mathbb{E}[\Delta(\mathcal{D}, S_0)]$.

We start by establishing a connection between the generalization error $\Delta$ and the spread of states $\mathcal{S}$. 
Following \cite{bartlett2002rademacher}, we can write the upper bound on the expected generalization gap, with high probability over the random draw of the training sample $\mathcal{S}$, by leveraging the Rademacher complexity of the hypothesis class $\mathcal{H}$:
\begin{equation} \label{eq:4}
\mathbb{E}[\Delta(\mathcal{D}, S)] \leq 2\, \mathfrak{\hat R}_n(\mathcal{H}; S)
\end{equation}
Here $\mathfrak{\hat R}_n(\mathcal{H}; S)$ denotes the Rademacher complexity with respect to the sample set $S = \{s_1, \dots, s_n\}$, which measures how well a hypothesis class $\mathcal{H}$ can fit random noise \cite{koltchinskii2002empirical}. 
For a linear hypothesis class $h_w(s) = \langle w, s \rangle$ with $\|w\|\leq B$, where $w$ is the parameter vector of the linear hypothesis, the Rademacher complexity can be expressed as \cite{lampertuseful}:
\begin{equation} \label{eq:5}
\widehat{\mathfrak{R}}_n(\mathcal{H}; S) \;=\; \frac{B}{n} \sqrt{\sum_{i=1}^n \|s^i\|^2}
\end{equation}

Building on our assumption that the distribution of $S$ is centered, we have $\mathbb{E}[S]=0$. 
After combining Equations~(\ref{eq:4}) and (\ref{eq:5}), we arrive at the following bound:
\begin{equation} \label{eq:6}
\mathbb{E}[\Delta(\mathcal{D}, S)] \;\leq\; O\!\left(\frac{B}{\sqrt{n}} \sqrt{\operatorname{tr}(\Sigma)} \right)
\end{equation}
\eq{6} provides an approximate upper bound on the generalization error conditioned on a constant norm bound $B$ of the weights, sample size $n$, and the covariance of the state distribution $\Sigma$.
This result suggests that the upper bound on the generalization gap \textit{scales as a function of the variance in the state distribution}. 
By learning a set of anchor states that reduces the variance in the state distribution, we can reduce the upper bound on the generalization gap and thus improve policy performance on unexpected initial states. 

We can also formalize this intuition by examining the state distribution and analyzing the generalization error bound from an information-theoretic perspective. Following \cite{kawaguchi2023does}, the generalization error can be upper bounded as:  
\begin{equation}
\tilde{\mathcal{O}}\left(\sqrt{\frac{I(S_0;S_b \mid \hat S_t)+1}{n}}\right) \quad \text{as}~ n \to \infty
\end{equation}
where $I$ is the information gain and $S_b$ is set of intermediate states with $0 < b < t$. 
Consider a reduction policy \(\pi^\prime\) that leads to a linear transition operator $T_{\pi^\prime, \theta}$ which maps \(S_b\) to anchor states, such that \(S_a \sim T_{\pi^\prime, \theta}(S_b)\). 
By the data processing inequality \cite{beaudry2011intuitive}, any operation on \(S_t\) yields  
\mbox{$I(S_0; S_a \mid \hat S_t) \leq I(S_0; S_b \mid \hat S_t)$}. Furthermore, if the transformation reduces the effective rank, i.e., \mbox{\(\mathrm{rank}(\mathrm{Cov}(S_a)) < d\)}, then necessarily \(\mathrm{rank}(T_{\pi^\prime, \theta}) < d\), making \(T_{\pi^\prime, \theta}\) non-invertible. In this case, provided that \(S_b\) is not already confined to a lower-dimensional subspace, we reach:  
\begin{equation}
I(S_0; S_a \mid \hat S_t) < I(S_0; S_b \mid \hat S_t)
\end{equation}
This provides an alternative analysis that demonstrates concentrating \(S_b\) reduces the information it carries about the input, thereby tightening the generalization error bound. 

\subsection{Reduction and Information-Preservation Trade-off}

In Section~\ref{AA} we prove that reducing the variance in the state distribution improves the generalization bounds.
However, this may also simultaneously reduce the mutual information between $S_b$ and the goal states $\hat S_t$ such that $I(S_a ; S_g) < I(S_b ; S_g)$, making it harder for the policy to learn mappings from intermediate states to goal states. 
Returning to our motivating example, restructuring the scene by placing the target at a same position behind the box reduces the spread of states, but the target remains invisible to the robot and cannot be reliably grasped.
In contrast, reconstructing the scene by removing the obstacle and exposing the target both reduces the spread of states and preserves the information necessary for the robot to execute the subsequent open-loop grasp.

The challenge therefore lies in constructing anchor states $S_a$ that (i) reduce the spread of $S_t$ to tighten the generalization error bound, while (ii) preserving the mutual information $I(S_a; S_g)$ so that $S_a$ remains sufficiently informative about the final state $S_g$.
In the next section we will present our method, ReSET, where we use action agnostic human videos as a source to create anchor states that intuitively satisfy both of these requirements.
\section{ReSET} \label{sec:method}

Building on our theoretical analysis, our algorithmic approach learns a reduction policy that rearranges the environment so that it is easier for the default policy to succeed.
More formally, the reduction policy takes the initial states to anchor states. 
By first compressing the trajectory into a narrower region of the state space, the reduction policy ensures that subsequent learning occurs on a distribution with smaller variance, which improves sample efficiency and increases the likelihood that the default policy succeeds. 
In this section we explain how the reduction policy is learned.

In order to reduce the state distribution into anchor states while preserving mutual information with the goal states, our method leverages \textit{action-agnostic human videos} as a natural source for constructing anchor states. 
We recognize that humans instinctively restructure environments in ways that simplify task execution, while also maintaining information relevant to the final goal (e.g., humans move the box so they can see the target cup). 
We aim to encode how humans simplify the initial states --- as captured by videos of human motion --- into our reduction policy.

To learn meaningful action plans from how humans restructure the environment, we draw inspiration from recent approaches that leverage action-agnostic videos for policy learning \cite{xu2024flow, wang2023mimicplay}, and represent human videos as point flows to extract abstract actions. 
Building on these representations, we train a flow-based reduction policy that imitates human strategies and simplifies the environment based on initial observations. 
Practically, we propose a learning framework consisting of three key components:
\begin{enumerate}[leftmargin=*]
\item A scoring network that evaluates the environment and determines if we should keep simplifying the environment or proceed to rollout the base policy (\fig{method} (a)). 
\item A flow generation network that predicts point flows which captures the intuitive strategies humans will employ to restructure the environment. (\fig{method} (b)).
\item A task-agnostic reduction policy, $\pi^\prime$. Given the flow proposed by the flow generation network, $\pi^{\prime}$ will achieve the desired outcome indicated by the flow (\fig{method} (c)).
\end{enumerate}
In what follows we will discuss each component in detail.

\subsection{Scoring Network}
The scoring network determines when to switch from the reduction policy to the base policy.
In other words, it distinguishes anchor states $S_a$ from other initial states. 
We introduce the \textit{scene score} $\mathcal{C}$ to quantify how likely it is for a trained policy to succeed under the initial configuration. 
Given a human video $O_h = \{o^t_h\}^T_{t=1}$, the score assigned to each frame $o_h^t$ is labeled with a fixed temporal prior; our assumption here is that over the course of the video the human is reconfiguring the environment into simpler state.
We model this prior using a monotonically decreasing function with respect to time $t$.
Frames closer to the end of the video receive lower values, indicating that the base policy is more likely to succeed by rolling out from those later frames. Specifically, we employ a second-degree polynomial in the temporal index $t$ with a negative leading coefficient:
\begin{equation} \label{eq:num9}
    \tilde{\mathcal{C}}_t = \alpha - \left(\frac{t}{\beta}\right)^2
\end{equation}
Here $\alpha$ and $\beta$ are tunable parameters controlling the scale and curvature of the decay. 
We found a parabolic decay helpful as it provides stronger discrimination at larger values of $t$, where the curvature increases with $t$.
Designers can modify \eq{num9} to other monotonically decreasing functions; our method is not tied to any specific scoring index.

The scoring network \mbox{$f : \mathcal{O}_h \longrightarrow \mathcal{C}$} is trained solely on human motion videos: it predicts the surrogate scene score of a given observation during rollout with the loss function:
\begin{equation}
    \mathcal{L}_{\text{score}} = \frac{1}{T} \sum_{t=1}^{T} \big( f(o^t) - \tilde{\mathcal C}_t \big)^2
\end{equation}
We then introduce a threshold on the scene score $\mathcal{\hat C}$ and compare it with the predicted score $f(o^t)$ to determine when to switch from the reduction policy to the base policy (see \fig{method} right). Returning to our motivating example, observations where the cup is directly visible receive a lower score $\mathcal{C}$, indicating that these states are more suitable for executing the base policy. 

\subsection{Flow Generation Network} \label{sec:BB}

\begin{figure}[t]
	\begin{center}
 		\includegraphics[width=\linewidth]{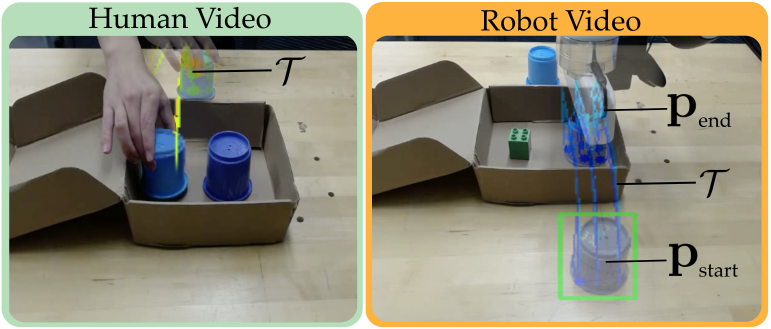}
		\caption{Our data preprocessing pipeline. \textit{Left}: we track the point flow of object movement $\mathcal{T}$ as a guidance to restructure the environment. \textit{Right}: For the robot videos, we locate the object by identifying points with large displacements. Alongside the point flow, we record the parameters of the associated action primitive, i.e., the start and end positions, as well as the rotations of the robot's end effector.} 
		\label{fig:data}
	\end{center}
    \vspace{-2.5em}
\end{figure}

\begin{figure*}[t]
	\begin{center}
 		\includegraphics[width=\linewidth]{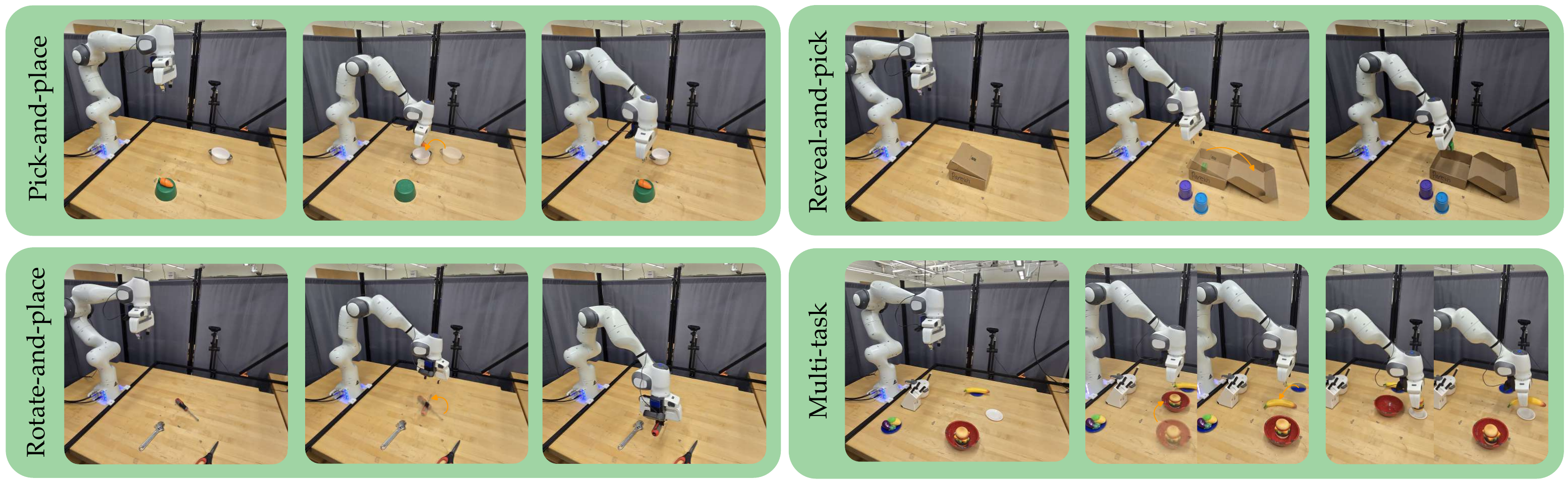}
		\caption{We evaluated ReSET on four real-world tasks. For each task, the first image illustrates a randomly initialized scene configuration, the second shows one of the anchor states reached after our reduction policy, and the third shows the subsequent execution of the base policy.}
		\label{fig:task}
	\end{center}
        \vspace{-2.5em}
\end{figure*}

We now have a way to determine when we need to rearrange the environment and when we are ready to execute the default policy. 
But how should our reduction policy actually change the environment? 
Here we draw from human videos.
Our approach attempts to transfer the actions performed by humans to the robot learner; i.e., we capture how humans prepare the environment, and enable robots to recognize and perform similar motions.
The heart of this challenge is mapping \textit{action-agnostic} human behaviors (i.e., humans manipulating objects with their own hands) into embodied actions the robot arm can perform.

We facilitate transfer across the embodiment gap by leveraging object \textit{point flows}.
We record the point flows of manipulated objects $\mathcal{T}$ in human videos using an off-the-shelf point tracking foundation model \textit{CoTracker3}~\cite{karaev24cotracker3} (see \fig{data}).
We then train a flow generation network \mbox{$g: \mathcal{O} \longrightarrow \mathcal{T}$} based on the videos and the extracted point flows.
Given an initial scene observation $o_0$, this network predicts a point flow that represents the most likely object movements consistent with how humans might have rearranged those objects to reduce the scene score $\mathcal{C}$.
To make point flows $\mathcal{T}$ easier to predict, we reduce the temporal resolution of the flow trajectories by performing linear interpolation on each point sequence along the temporal dimension. 
In particular, we down-sample the trajectories to a fixed horizon of $18$ time steps, thereby preserving the continuous temporal structure of the original data while maintaining expressiveness. 

In order to encode a richer representation of the restructuring actions, we seek to model the spatio-temporal dependencies between points.
We adopt a spatiotemporal transformer architecture inspired by~\cite{bruce2024genie} which interleaves spatial and temporal attention layers (see \fig{method}).
The input image is first encoded into patch tokens using a pretrained \textit{DINOv2} encoder~\cite{oquab2023dinov2}, yielding a representation of shape $1 \times 400 \times 768$. 
These tokens are reconstructed by applying another patching along the spatial dimension with a $5 \times 5$ grid, resulting in $P = 16$ patches, each with token dimension $D = 5 \times 5 \times 768$.
The patches are then repeated across $T$ steps, corresponding to the horizon of the point flow to be predicted. 
This produces a feature sequence of shape $T \times P \times D$, which is then passed into the flow generator. 
Notably, we observe that incorporating a learnable temporal encoding along the time dimension enhances the model's capacity to capture richer point flow representations.

Note that the flow generation network functions as a planner for restructuring the scene and is not restricted to the point-flow representation in our method. 
In our motivating example, its purpose is to indicate how the obstructor should be repositioned to reveal the target object behind it. 
Any approach capable of emulating human-like reasoning could serve the same role, such as a vision–language model. 

\subsection{Task-Agnostic Flow-Based Reduction Policy}

The flow generation network from Section~\ref{sec:BB} provides guidance on how the scene should be manipulated to reach anchor states and make the task easier to complete.
Our final step is for the robot to actually perform these actions and restructure the environment.
This is our \textit{reduction policy}: a mapping from images and predicted point flows into robot behaviors.
Our reduction policy is trained using task-agnostic teleoperated robot data (e.g., videos of the robot playing with objects in the environment).
In theory, the reduction policy could map to low-level robot actions.
But because it remains challenging to directly predict fine-grained actions based on a point flow, our reduction policy instead outputs the parameters for a set of predefined action primitives. 
Across all setups, we observe that reconstruction actions can be grouped into three categories: (i) pick-and-place, (ii) push-and-pull, and (iii) rotation.
Hence, we need primitives that can handle these sorts of behaviors.

We represent the action primitive space as $\mathcal{A} = \{(c, \mathbf{p})\}$, where \(c \in \{c_1,c_2,c_3\}\) denotes the primitive class (pick-and-place, push-and-pull, rotate), and \(\mathbf{p}\) are the continuous parameters associated with each primitive.
For example, in the pick-and-place primitives \(\mathbf{p}\) are the coordinates for the pick and the place.
Practically, we first get the bounding box of the manipulated object around the points with the largest displacement in the scene. 
The action primitives are then parsed by a temporal window between the beginning and end of the detected motion (see \fig{data}).

The reduction policy takes as input a point flow $\mathcal{T}$ and a robot initial observation $O_r^0$ and predicts an action primitive $\mathcal{A}$, $\pi^{\prime}: \; \mathcal{T} \times \mathcal{O}^0_r \longrightarrow\; \mathcal{A}$. 
We use a transformer architecture as backbone. 
A learnable action token is first initialized and concatenated with the input tokens, consisting of the point flow and robot observation features. 
This joint token sequence is then processed by the transformer to produce the predicted action primitive $\mathcal{A}$. 
To train $\pi^\prime$ we use a composite loss that combines classification loss for the primitive type and regression loss for its parameters. Given model predictions \((\hat{c}, \hat{\mathbf{p}})\), the total loss is defined as:
\begin{equation}
\mathcal{L_{\pi^\prime}} = \lambda_{\text{cls}} \, \mathcal{L}_{\text{cls}}(c, \hat{c}) \;+\; \lambda_{\text{reg}} \, \mathcal{L}_{\text{reg}}(\mathbf{p}, \hat{\mathbf{p}}),
\end{equation}
where \((c, \mathbf{p})\) denote the ground-truth primitive type and parameters. The first term, 
\(
\mathcal{L}_{\text{cls}}(c, \hat{c}) 
\)
is the cross-entropy loss over primitive classes, and
\(
\mathcal{L}_{\text{reg}}(\mathbf{p}, \hat{\mathbf{p}})
\)
is the mean-squared error (MSE) loss for parameters. The coefficients \(\lambda_{\text{cls}}, \lambda_{\text{reg}} > 0\) balance the two objectives. 

\section{Experiments} \label{sec:experiment}

\begin{figure*}[t]
	\begin{center}
 		\includegraphics[width=0.9\linewidth]{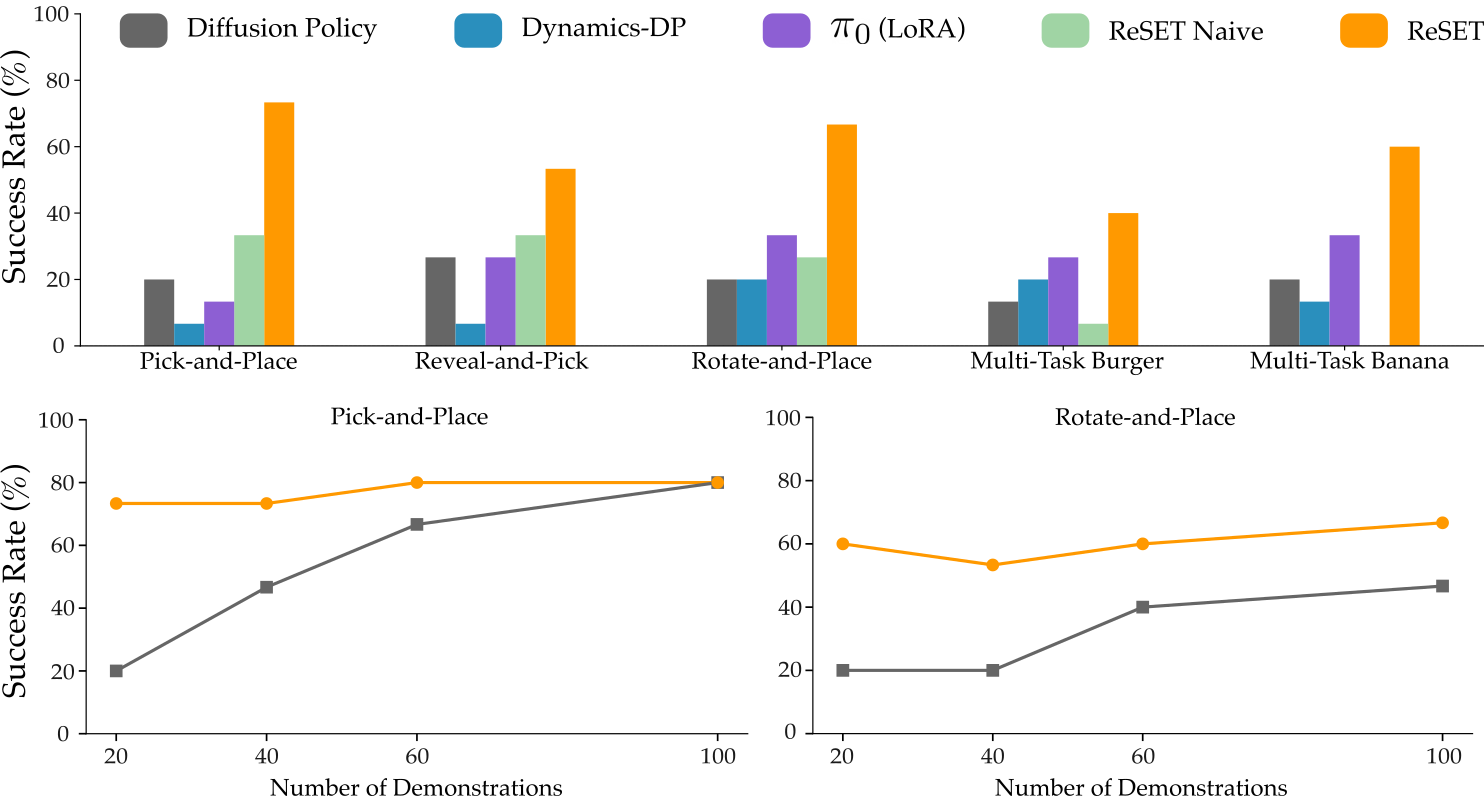}
		\caption{Results from our real-world experiments. Each task was evaluated on $15$ test scenarios, including out-of-distribution states that were not covered in the expert robot demonstrations. \textit{Top}: Success rate over four tasks. \textit{Bottom}: Performance of ReSET and Diffusion Policy with $20$, $40$, $60$, and $100$ demonstrations. We observe that across all tasks ReSET achieves a higher success rate and outperforms the baselines. We also observe that --- across different amounts of data available for training --- ReSET achieves a higher success rate as compared to Diffusion Policy.}
		\label{fig:success}
	\end{center}
        \vspace{-2.5em}
\end{figure*}

We experimentally compare ReSET to state-of-the-art alternatives. 
Within our experiments we evaluate tabletop object manipulation tasks that require composite, long-horizon reasoning by a robot arm. 
We focus on the interplay between training data and task performance, aiming to answer the following question: does introducing a reduction policy that restructures the environment reduce the effort of data collection --- both for training the reduction policy itself and for the base model? For demonstrations of our experiments, please see the \href{https://reset2025paper.github.io}{accompanying videos}.

\p{Baselines} We compare ReSET with multiple visual imitation learning baselines on a variety of task settings. 

    \subsubsection{ReSET Naive}
    An ablation of ReSET. Rather than learning a flow-based reduction policy, ReSET Naive directly maps from the initial observation to corrective actions. The policy is trained on an augmented dataset that explicitly aligns initial observations from human demonstrations and actions from robot play data based on the most similar flows, measured via the $\| \circ \|_2$ norm of the flow difference. Comparing to this baseline will show that our flow-based reduction policy is a necessary component for learning a more diverse range of reconstructive actions.
    
    \subsubsection{Diffusion Policy~\cite{chi2024diffusionpolicy}}
    Uses same rollout policy as our method, but does not call ReSET to prepare the environment.
    
    \subsubsection{Dynamics-DP~\cite{wu2025neural}} Generates augmented data on top of expert demonstrations to train a more robust diffusion policy. We implement the method introduced in the original paper and adapted an action decoder proposed 
    by \cite{gao2025adaworld}, extending it into a visual dynamics model for Model Predictive Path Integral (MPPI) \cite{doi:10.2514/1.G001921} Planning.
    
    \subsubsection{$\mathbf{\pi_0}$~\cite{black2024pi_0}}
    A VLA baseline. The policy leverage large-scale multimodal training data to align visual perception with language understanding. We take the flow matching version of the $\pi_0$ policy, and perform low-memory finetuning (LoRA) \cite{hu2022lora} with task specific data for $30,000$ steps.

\p{Setup} Our experiments are conducted using a Franka Emika robot arm on a tabletop manipulation setup. 
We employ a GELLO controller \cite{wu2024gello} to teleoperate the robot, collecting both the play dataset used for training the reduction policy and the task-oriented expert dataset used for training the base policy. The experimental setup incorporates two cameras. The static side-view camera is used as a primary camera for all datasets, and the gripper-mounted camera is used as a secondary camera in the demonstration dataset.

\p{Tasks}
We evaluated ReSET against the baselines on four real-world tasks (see \fig{task}).
\begin{enumerate}[leftmargin=*]
    \item \textit{Pick-and-place}: Grasp the carrot from its initial location and deposit it into the bowl.
    \item \textit{Reveal-and-pick}: Actively remove obstructing objects to uncover the hidden block, then grasp that block.
    \item \textit{Rotate-and-place}: Pick up a screwdriver that is originally at an angle, and place it parallel to the other tools.
    \item \textit{Multi-task}: Serve either a burger or a banana based on user input. The two foods are on separate plates.
\end{enumerate}

\p{Results}
\fig{success} compares the performance of each method across the four tasks. 
All methods are trained on $20$ expert robot demonstrations, with the exception of Dynamics-DP, which also uses an augmented dataset.
We collected the same amount of augmented data as human data for fair comparison. 
ReSET and its naive point-track matching variant incorporate $20$ action-agnostic human videos and $20$ minutes of robot play data for each task.
We examine the data efficiency of our approach in the following sections. 
Also note that ReSET combines robot play data for all the tasks and learns a single reduction policy. Each task was evaluated on $15$ test scenarios, approximately $80\%$ of which featured distributional shifts relative to the training set. 

\subsection{How well does ReSET recover from unexpected states?}

We found that ReSET learns the restructuring strategies featured in the human videos and executes the reduction policy from out-of-distribution states.
For example, in the pick-and-place task, the robot pulls the white bowl towards the carrot so that it is easier for the base policy to accurately drop the carrot into the bowl. 
In the rotate-and-place task, the robot first rotates the screwdriver into an easier-to-grasp orientation before picking it up and placing it among other tools (see \fig{task} and supplemental videos). 

In long horizon tasks like reveal-and-pick, ReSET is able to execute a sequence of reconstructions of the scene until an anchor state is reached. 
If the robot observes the block is still covered under a cup after opening the box, it will proceed to pick up the cup and reveal the block. 
In a multi-task setup, ReSET generates different restructuring actions according to different instructions (``Serve burger'' or ``Serve orange''). 

\subsection{How does a flow-based reduction policy help?}

Comparing ReSET to ReSET Naive underscores the importance of incorporating flow into the reduction model.
ReSET adapts more effectively to variations in the initial observation, generating richer restructuring plans than the naive ablation.
Indeed, point track matching consistently fails to handle slight positional shifts of objects. 
This behavior is expected: during training for ReSET Naive the same robot action is often aligned with multiple distinct initial observations from the human dataset, thereby reducing the effective diversity of action variants available for learning. 
Our flow based reduction policy captures more diverse actions from robot play data, enabling more flexible and adaptive behavior.

\subsection{Is ReSET more data efficient that the alternatives?}

To demonstrate that our method remains data efficient despite its reliance on both human and robot play data, we trained diffusion policies with an increased number of expert robot demonstrations and compared their performance against ReSET (\fig{success} bottom).
We found that diffusion policies requires at least $70$ expert demonstrations for each task to reach comparable performance.
Based on our experience, collecting this amount of data takes at least one hour for an expert demonstrator. In contrast, ReSET only requires about $10$ minutes of task-specific human corrections plus $20$ minutes of task-invariant robot play data.
This highlights that with ReSET the robot can recover from out-of-distribution scenarios with substantially less time spent collecting demonstrations.

\section{Conclusion} \label{sec:conclusion}

In this paper we introduced ReSET, an algorithmic framework that restructures environments for easier task execution.
ReSET learns from two sources of training data: action-agnostic videos of humans rearranging the scene, and task-agnostic videos of teleoperated robot play.
From the human videos we extract i) a score that determines whether the robot should roll-out its base policy or simplify the environment, and ii) a flow prediction that captures how humans would manipulate objects.
Using the robot play videos, we then convert these into iii) a policy to map the point flow into robot actions that interact with the environment and lower the difficulty score.
Overall, our learned approach rearranges the environment to bring out-of-distribution initial states back into a smaller, known distribution (i.e., anchor states).
Our theoretical analysis suggests that moving to anchor states before executing the baseline policy improves generalization in a data-efficient manner.
This analysis is supported by our experiments, where we show that ReSET leads to higher performance across multiple tasks and data amounts.

%%%%%%%%%%%%%%%%%%%%%%%%%%%%%%%%%%%%%%%%%%%%%%%%%%%%%%%%%%%%%%%%%%%%%%%%%%%%%%%%%

% \newpage
\balance
\bibliographystyle{IEEEtran}
\bibliography{IEEEabrv,bibtex}

\end{document}